\pdfoutput=1

\documentclass[11pt]{article}

\usepackage[final]{acl}

\usepackage{times}
\usepackage{latexsym}
\usepackage{todonotes}

\usepackage[T1]{fontenc}

\usepackage[utf8]{inputenc}
\usepackage{float}
\usepackage{listings}
\usepackage{microtype}

\usepackage{inconsolata}

\usepackage{graphicx}

\usepackage{booktabs}
\usepackage{ragged2e} 
\usepackage{multirow}

\usepackage{amsmath}

\newcommand{\fig}[3][0.9]{
\begin{figure}
\centering
\includegraphics[width=#1\columnwidth]{images/#2}
        \caption{#3}~\label{fig:#2}
\end{figure}}

\newcommand{\ACRO}[1]{\textsc{#1}}
\newcommand{\AMR}{\ACRO{amr}}
\newcommand{\ARRAU}{\ACRO{arrau}}
\newcommand{\CODICRAC}{\ACRO{codi}/\ACRO{crac}}
\newcommand{\LINGEX}[1]{\textit{#1}}
\newcommand{\MDC}{\ACRO{mdc}}
\newcommand{\MDCR}{\ACRO{mdc-r}}
\newcommand{\MMAX}{\ACRO{mmax}2}
\newcommand{\REC}{\ACRO{rec}}

\title{MDC-R: The Minecraft Dialogue Corpus with Reference}

 \author{
     \textbf{Chris Madge\textsuperscript{1}},
     \textbf{Maris Camilleri\textsuperscript{1}},
     \textbf{Paloma Carretero Garcia\textsuperscript{1}},
     \textbf{Vanja Karan\textsuperscript{2}},
     \\
     \textbf{Juexi Shao\textsuperscript{1}},
     \textbf{Prashant Jayannavar\textsuperscript{3}},
     \textbf{Julian Hough\textsuperscript{4}},
     \textbf{Benjamin Roth\textsuperscript{2}},
     \textbf{Massimo Poesio\textsuperscript{1}}
     \\
     \textsuperscript{1}Queen Mary University of London,
     \textsuperscript{2}Universit{\"{a}}t Wien,\\
     \textsuperscript{3}University of Illinois,
     \textsuperscript{4}Swansea University
     \\
     \small{\textbf{Correspondence:} \href{mailto:m.poesio@qmul.ac.uk}{m.poesio@qmul.ac.uk}}
 }

\begin{document}
\maketitle
\begin{abstract}
We introduce the Minecraft Dialogue Corpus with Reference ({\MDCR}). {\MDCR} is a new language resource that supplements the original Minecraft Dialogue Corpus ({\MDC}) with expert  annotations of anaphoric and deictic reference. 
{\MDC}'s task-orientated, multi-turn, situated dialogue in a dynamic environment has motivated multiple annotation efforts, owing to the interesting linguistic phenomena that this setting gives rise to. 
We believe it can serve as a valuable resource when annotated with reference, too.
Here, we discuss our method of annotation and the resulting corpus, and provide both a quantitative and a qualitative analysis of the data. Furthermore, we carry out a short experiment demonstrating the usefulness of our corpus for referring expression comprehension.
\end{abstract}

\section{Introduction}

Dialogue games are games in which conversational agents impersonating characters can learn to perform tasks or improve their communicative ability by interacting with human players or other artificial agents. Such games offer an exciting opportunity to study how conversational agents can carry out interaction \cite{johnson2016malmo,urbanek2019learning,narayan2019collaborative,szlam2019build,bara-et-al:EMNLP21,kiseleva2022interactive}. 
Virtual world dialogue games in particular may approach the complexity of the real world \cite{shridhar2020alfred,puig2018virtualhome,kolve2017ai2} and virtual agents operating in such virtual worlds need to be able to develop a variety of interactional skills to be perceived as ``real'' .
Dialogue games have 
been argued to be the best benchmark for (grounded) spoken language understanding \cite{schlangen2023dialogue}.  
Games are also engaging, so human participants are more likely to be motivated 
\cite{von2006games,chamberlain2008phrase,poesio-et-al:ACMTIIS,poesio-et-al:NAACL19,szlam2019build}.

Virtual world games are a particularly promising domain to study reference to entities in a visual situation \cite{johnson2017clevr,qi2020reverie,islam2022caesar} and anaphoric reference in dialogue \cite{yu2022codi}.
Most research on referring expression comprehension has focused on static images 
and most research on anaphoric reference has focused on written text \cite{poesio-et-al:ARL23}, 
but 
dialogue presents unique challenges for both anaphoric and deictic reference, as reference in dialogue is more fluid. 
Participants propose, negotiate and if necessary repair their common understanding of referents \cite{clark1986referring}.
Some recent efforts have looked at  reference  in situated dialogue  \cite{loaiciga2022anaphoric,gigliobianco2024learning}, but this research focuses on static environments, rather than a dynamic one in which one of the interlocutors may manipulate the environment over the course of the dialogue.

In this work, we introduce a corpus of conversations between human agents carrying out tasks in the virtual  world Minecraft Collaborative Dialogue Task \cite{narayan2019collaborative}. This task sees two parties collaborate to build a structure in a virtual  world,
annotated for deictic and anaphoric reference.
This yields a dataset that poses unique annotation challenges and opportunities, given its dynamic environment that continuously changes throughout the conversation as it is manipulated by the human participants, introducing complex grounding, misunderstandings and ambiguity. 
The voxel world task-orientated dialogue requires addressing abstract shapes composed of blocks, and compositions of those abstract shapes. This results in another interesting phenomena known as conceptual pacts \cite{brennan1996conceptual}, in which collaborating participants in a conversation come to an agreement over terms referring to entities. 


The contributions of this paper are as follows: 
1. a new approach to annotating reference in dynamically changing visual worlds;
2. a new annotation for deictic and anaphoric reference of the popular Minecraft Dialogue Corpus \footnote{Freely available from \url{https://github.com/arciduca-project/MDC-R}};
3. an analysis of the issues with reference arising;
and
4. a baseline model for reference resolution in this corpus.

The rest of the paper is structured as follows. 
We 
 review 
 related work in Section~\ref{sec:relwork}, followed by a discussion of the base {\MDC} dataset in Section~\ref{sec:mdc} and \ref{sec:ref-in-mdc}. Next, Section~\ref{sec:annotation} outlines our annotation methodology. We then describe our prediction experiments and provide corresponding discussion in Section~\ref{sec:obs} and Section~\ref{sec:experiment}, respectively. 

\section{Related Work}\label{sec:relwork}

\subsection{Virtual world dialogue games}

Aside from the Minecraft Collaborative Building Task \cite{narayan2019collaborative}, discussed 
in Section \ref{sec:mdc}, there have been  a few other works that  use  virtual world games to collect task orientated dialogue between human interlocutors. 

\citet{ogawa2020gamification} created a platform 
for collection of situated dialogue as an extension to \textit{Minecraft}.  As part of their work, they propose a scenario inspired by the \textit{Map Task} \cite{anderson1991hcrc}, in a \textit{Minecraft} setting. 
\textit{Mindcraft} is a \textit{Minecraft}-like game in which two players, with differing skillsets collaborate to complete a shared goal,
communicating to form a common understanding of their respective skills in relation to the given mission \cite{bara-et-al:EMNLP21}.  The aim of the work is to understand how players establish a mental model in a situated task, and how this is affected by communication and the shared environment.

In the \textit{CerealBar} game 
\cite{suhr2019executing} 
two separate uniquely constrained participants, must cooperate to collect cards in a 3D environment. The leader, possessing greater observability but comparatively reduced movement, instructs the follower (who is unable to respond). 

Virtual world games for 
dialogue are not limited to 3D interfaces.
\textit{LIGHT} \cite{urbanek2019learning} is a text-based multiplayer adventure-fantasy world, designed for studying grounded dialogue in a situated environment. Participants are able to both speak and act in text form.  
A corpus of 11k episodes was collected, some of which have been used in reference-based tasks \cite{yu2022codi}.


\vspace{-4pt}
\subsection{Referring Expression Comprehension}

In a conversation in which the participants share a visual scene two possible types of reference are possible: \textit{anaphoric} reference to entities introduced in the language, and \textit{deictic} reference to  objects in the visual scene that may or may not have been mentioned before. 
Conversational agents interacting in a virtual world need both the ability to comprehend referring expressions and to generate them \cite{van-deemter:book:referring}.

\paragraph{The REC task}
In recent years, 
Referring Expression Comprehension ({\REC}) \cite{kazemzadeh-etal-2014-referitgame,mao2016generation,nagaraja2016modeling} has become a fundamental task in vision-and-language research.
In this field, it is typically defined as   identifying the correct bounding box for a target object described by a free-form text. Over the years, the scope of {\REC} has expanded to include various types of textual inputs --- ranging from conversational queries \cite{de2017guesswhat} and scene-aware contexts \cite{chen2023advancing} to knowledge-based cues \cite{wang2020give,you-etal-2022-find} --- which collectively challenge multimodal models to harness more advanced reasoning skills. In parallel, researchers have leveraged simulated environments \cite{johnson2017clevr,qi2020reverie,islam2022caesar} to study compositional and embodied reasoning in settings where objects are meticulously customised. Despite the flexibility of these virtual worlds, existing {\REC} studies have yet to explore simulated settings enriched with extended, dialogue-centric text, highlighting a significant gap that our work aims to address.


\paragraph{Generalized REC}
Generalized Referring Expression Comprehension (\ACRO{grec})~\cite{GRES,GREC} extends traditional {\REC} by allowing expressions to refer to multiple or no targets, thereby enhancing its applicability in real-world scenarios. Recently, RECANTFormer~\cite{hemanthage2024recantformer} has employed a one-stage method to improve \ACRO{grec} recognition. Our corpus aligns with the \ACRO{grec} framework by providing bounding boxes for multiple targets within a single image and unique annotations for each target, as detailed in Section 5.1.





\paragraph{Visual Coreference}
In parallel with the {\REC} work,  
recent research in 
multimodal coreference resolution have underscored the significance of integrating visual cues into language understanding. For instance, \cite{goel2023referring} and \cite{goel2023semi} explore coreference resolution in image narrations, demonstrating that grounding textual entities in specific visual regions can substantially enhance reference disambiguation. In addition, research in visual dialogue has made notable progress with works such as \cite{yu2019see} and \cite{yu2022vd}, which leverage visual context to resolve pronoun references and thereby improve dialogue coherence and interpretability. \newcite{loaiciga2021annotating} delve into the annotation of anaphoric phenomena within situated dialogue, emphasizing how contextual information is critical for resolving referential ambiguities.


\vspace{-4pt}
\subsection{Datasets for studying reference in dialogue}


REX-J \cite{spanger2012rex} task sees two participants collaborating in separate roles 
on a tangram puzzle. The solver role communicates with the operator role, proposing a solution, while the operator manipulates the tangram pieces. 
The PentoRef task 
\cite{zarriess-etal-2016-pentoref}
also makes use of two human participants, an instruction giver (who can see the target structure), and an instruction follower (can manipulate pieces), collaborating to reproduce the target structure. The followers actions are used to infer the effectiveness of the instruction giver's utterances. Reference annotation on the resultant dataset is carried out by experts \cite{zarriess-etal-2016-pentoref}.
The corpus of \newcite{loaiciga2021annotating} features the \textit{Cups} task, in which two participants collaborate to identify discrepancies in an almost identical 3D environment, given images taken from two different perspectives. The dataset is expert annotated for reference with {\MMAX} \cite{muller2006multi}, according to ARRAU \cite{poesio-et-al:ARRAU3:manual} (similar to the approach used in this work).

All of the aforementioned datasets and methods of gathering dialogue promote interesting referential phenomena, including deictic reference and spatial relations with the two party instruction giver/follower paradigm being a popular format for soliciting this type of dialogue. In some respects, our offering could be seen as a combination and natural extension of the properties of prior corpora, in that we provide annotations over a dataset that incorporates a collaborative task-orientated dialogue, with action aligned utterances, and changing perspectives in a dynamic 3D situated environment.

\section{The Minecraft Dialogue Corpus 
}\label{sec:mdc}

\subsection{The Data}

The Minecraft Dialogue Corpus \cite{narayan2019collaborative} is a collection of conversations among human participants performing the Minecraft Collaborative Building Task, illustrated in 
Figure \ref{fig:mc}.
\fig{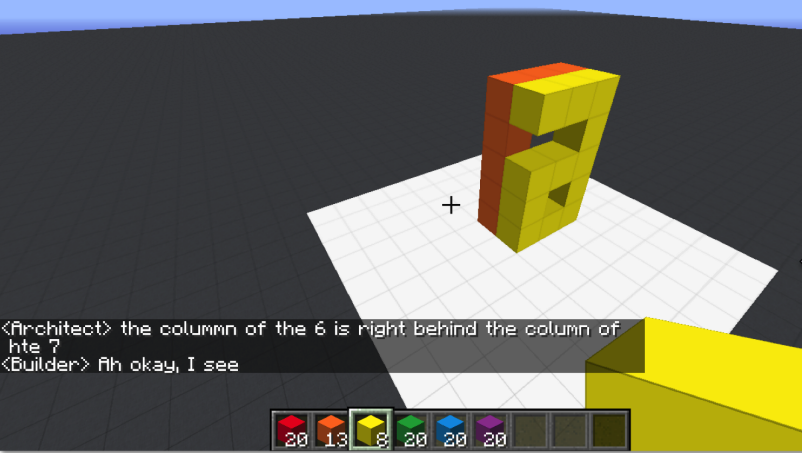}{Example of Minecraft Collaborative Builder Task from \cite{narayan2019collaborative}} 
In these conversations, two humans, \textit{the Architect} -- giving instructions and \emph{the Builder} -- executing them, cooperate to replicate a 3D structure in a 3D voxel based $11 \times 9 \times 11$ Cartesian coordinate based Minecraft world, with blocks of 6 different colours, accessed through  Malmo \cite{johnson2016malmo}.
The Architect  has full observability over both the target environment and the builders actions, but may not directly effect change to the environment, only converse with the builder.
The Builder is not able to see the target structure, but has an avatar that they can use to navigate and manipulate the environment with the goal of constructing the target structure.
This motivates a multi turn situated dialogue exhibiting complex phenomena, including references to actions, objects and abstract shapes.  
But also sees these develop and sometimes change as the target structure is created. Figure \ref{fig:mc} shows an example of this from the Builder's perspective.
No conversational constraints are imposed; bidirectional communication, clarification etc. are all permissible. 509 dialogues were collected, with an average length of 30 utterances, for 15,926 utterances total, and 113,116 tokens.


\subsection{Annotations}

The richness of linguistic phenomena observed in the {\MDC} has motivated several previous efforts to extend {\MDC} with various forms of annotation.  

\citet{bonn2020spatial} 
annotated 185 of the dialogues with an extended form of Abstract Meaning Representation ({\AMR}) \cite{banarescu2013abstract} that incorporates spatial relations.  
\citet{bonial2021builder} apply a separate extension of {\AMR} on {\MDC}, with the Dialogue-{\AMR} \cite{bonial2020dialogue} representation, a form of {\AMR} that captures the illocutionary force of dialogue acts.

\citet{thompson2024discourse} produced the  Minecraft Structured Dialogue Corpus (\ACRO{msdc}), that, through Segmented Discourse Representation Theory (\ACRO{sdrt}) \cite{asher2003logics,lascarides2007segmented}, gives a representation of the logical forms of the discourse in {\MDC}, connecting dialogue and agent actions to deliver a macrostructure that links narrative arcs and discourse relations (e.g. corrections, confirmations, acknowledgements etc.).

\vspace{-4pt}
\section{Reference in the 
MDC 
}\label{sec:ref-in-mdc}

\begin{figure}
\centering
\includegraphics[width=\columnwidth]{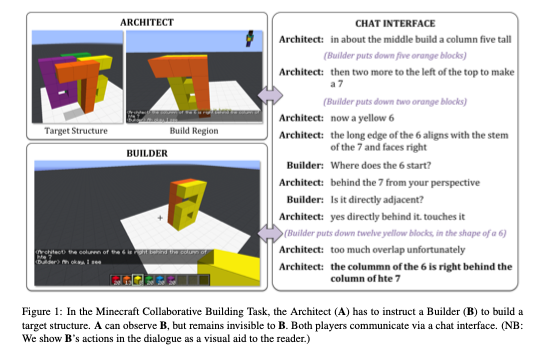}
\caption{References in a {\MDC} dialogue \cite{narayan2019collaborative} \label{fig:mdc:ref}}
\end{figure}

The {\MDC} dialogues contain a rich variety of examples of reference in dialogue, as illustrated by the dialogue in Figure \ref{fig:mdc:ref}.

The dialogues contain 
plenty of deictic references to the visual situation (\LINGEX{the 6}, \LINGEX{the 7}), many of which are bridging references (e.g., \LINGEX{the middle} in the first utterance, \LINGEX{the top}). 
But perhaps the most distinctive aspect of these conversations is that 
the visual situation dynamically changes throughout the dialogues as new objects are built. 
For instance, the object that can be described as \LINGEX{the 7} is the result of the Builder's actions of putting down five orange blocks, then adding two orange blocks, as per Architect instructions. 

Another distinctive feature of the corpus is that  the objects that emerge as a result of the Builder actions can be referred to using their distinctive shapes.  
In Figure \ref{fig:mdc:ref}, the Builder first builds an object that looks like a 7, and can therefore be referred to as \LINGEX{the 7}, then an object that can be referred to as \LINGEX{the 6}. 
In Figure \ref{fig:labels2}, the  overall object can be referred as \LINGEX{the arch}, and the set of blocks in the middle can be referred to as \LINGEX{the bell}, even if those terms have not been previously introduced.


\section{Our Annotation}\label{sec:annotation}

\subsection{Annotation Scheme}

The annotation scheme for {\MDCR} is based on the scheme developed for the {\CODICRAC} 2021 and 2022 shared tasks on anaphoric reference in dialogue \cite{yu2022codi}, which in turn is an extension of the {\ARRAU} annotation scheme \cite{poesio-et-al:ARRAU3:manual,poesio-et-al:CODI24:ARRAU3} to cover more dialogue phenomena. 
For {\MDCR}, we extended the {\CODICRAC} scheme to cover more complex
reference phenomena, building on
the proposals in \cite{loaiciga2022anaphoric}.
Reference to objects in a visual situation
could be done with strong reliability in the \ACRO{trains} portion of the {\ARRAU} corpus \cite{uryupina2020annotating}, in which the map was shown in an image separate from the annotation tool.
\cite{loaiciga2022anaphoric} obtained a Krippendorff $\alpha$ value of $0.55$ for the approach to annotation of references to the visual situation in which the image is shown on demand.  

\begin{figure}
\centering
\includegraphics[width=\columnwidth]{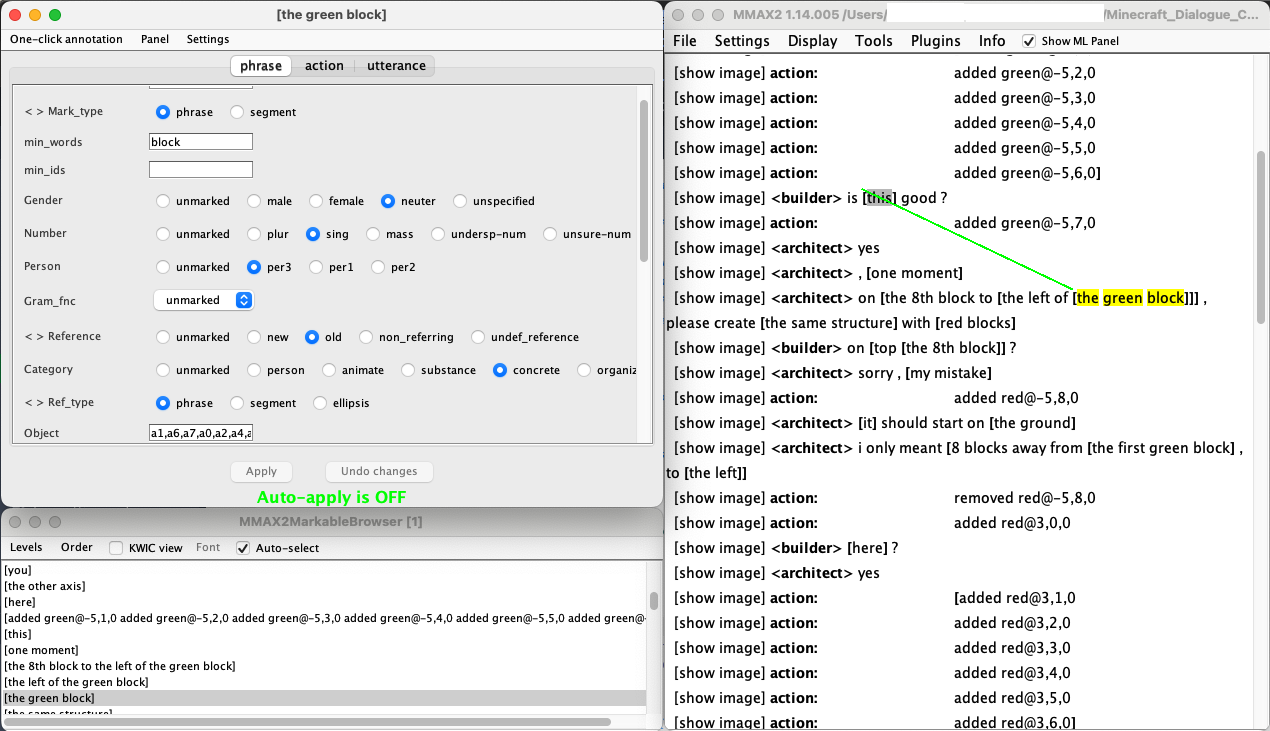}
\caption{Reference annotation and grounding in MMAX2~\label{fig:mmax2:before}}
\end{figure}

\begin{figure}
\centering
\includegraphics[width=\columnwidth]{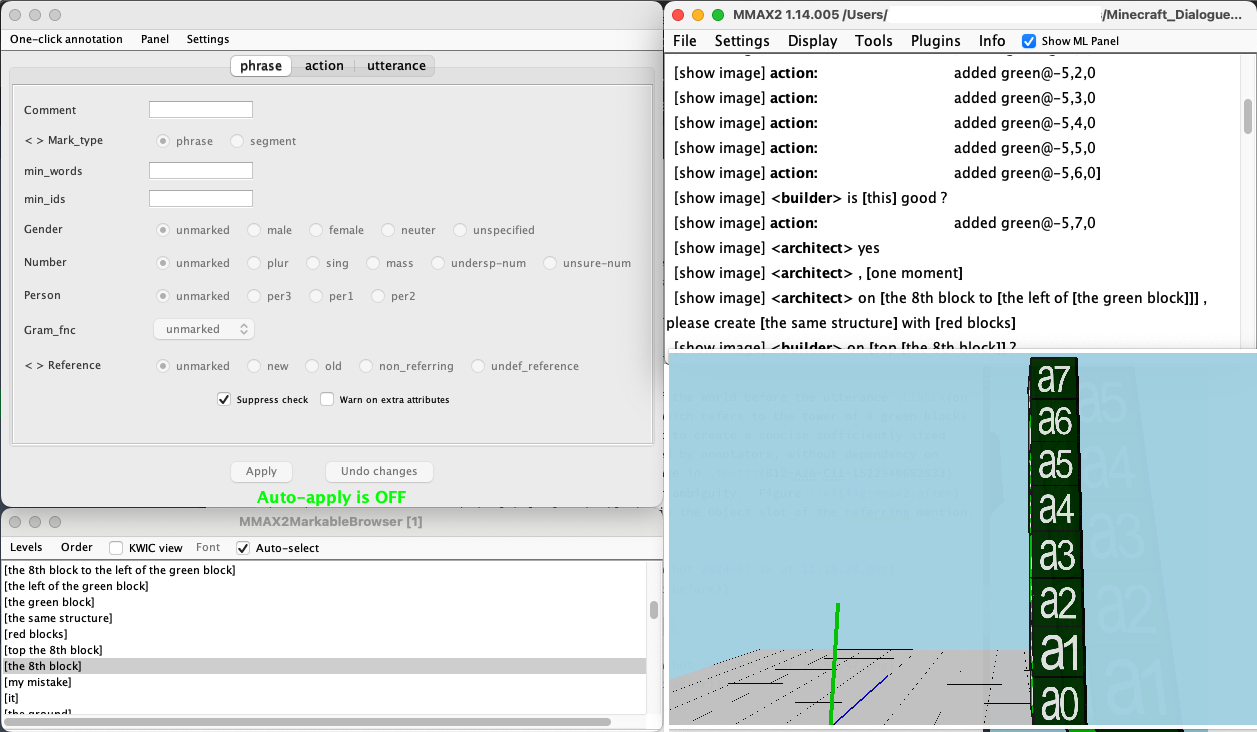}
\caption{Annotating references in MMAX2~\label{fig:mmax2:after}}
\end{figure}

The key challenges we had to tackle were how to specify references in the visual scene, and how to handle the constantly changing  dynamic environment. 
To address this challenge, we 
made 
some 
changes to the annotation tool {\MMAX} \cite{muller2006multi}. Similar to \cite{loaiciga2022anaphoric}, we support viewing an object labelled image of the current state of the world before each utterance, as seen by the builder.
In our adaptation of the proposal by Loaiciga et al. to the Minecraft world situation, 
each block is given a unique alphanumerical tuple label,
as 
shown in Figure \ref{fig:labels2}.
\fig[1]{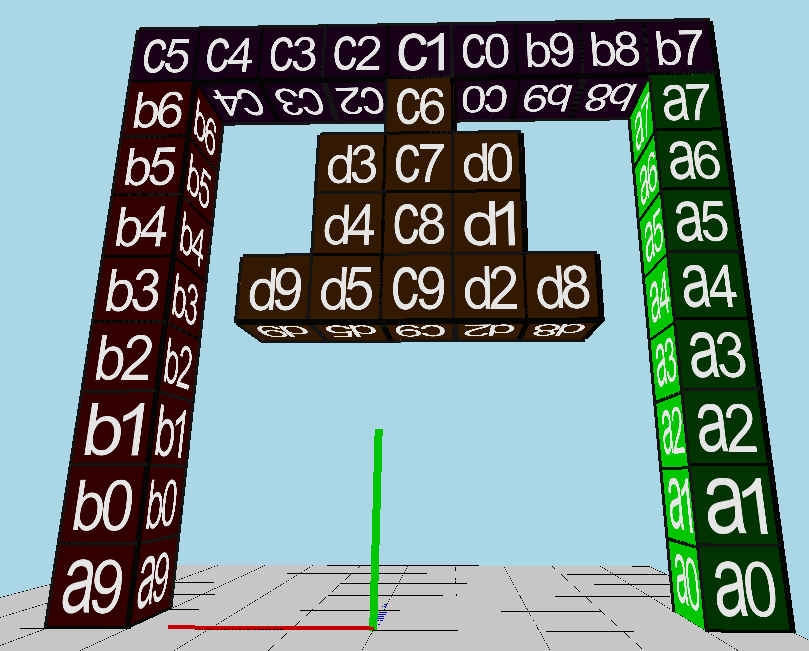}{Labelling {\MDC} with alphanumerical tuple indexes}
The annotators then enter the labels of the blocks referred to by the noun phrase in the `Object' slot.
Figures \ref{fig:mmax2:before} and  \ref{fig:mmax2:after} show how references to such states are annotated. 
In Figure \ref{fig:mmax2:before} we can see the state of the world before the utterance \LINGEX{on the block to the left of \underline{the green block}}, which refers to the tower of 8 green blocks a0,a1,a2,a3,a4,a5,a6,a7. We choose this labelling format to create a concise sufficiently sized address space that could be easily identified and entered by annotators, without dependency on colour (which can sometimes be misjudged e.g. brown/orange in \texttt{B12-A26-C11-1522940682033} \cite{narayan2019collaborative}), or introducing further ambiguity.  Figure \ref{fig:mmax2:after} shows how annotators entered the labels for the blocks in the Object slot of the referring mention.
A reliability 
test with this revised scheme and guidelines (2 experienced annotators double-coded 5 dialogues containing a total of about 400 markables), resulting in a preliminary $\kappa=0.43$. 
We expect this result to be a lower  bound as many of the disagreements are due to the coders identifying slightly different sets of blocks. 

\subsection{Bounding Boxes}

In the process of generating these labels, each label is tied to the 3D Cartesian coordinates of the block it is assigned to. 
This allows us to include a 2D bounding box that identifies the bounds of the block in the perspective based image. 
This makes {\MDCR} a versatile corpus, suitable both for versions of the {\REC} task in which systems have to output labels, and for the more traditional version of the {\REC} task in which systems have to output a bounding box (see Section \ref{sec:experiment} for an example of how {\MDCR} can be used in this way).


These images highlight one of the fundamental properties of this corpus: that reference to objects in the scene translates to reference to \emph{sets} of blocks that constitute these objects. 
This property means that a generalized notion of reference is required--one in which it is possible to refer not only to objects, but to sets of objects \cite{he-et-al:arxiv23:-grec-generalized-referring-expression,hemanthage2024recantformer}.

\subsection{Annotation and Statistics}
We selected at random a subset of 100  dialogues from {\MDC} and converted them into a format suitable for labelling in {\MMAX}. 
The data was annotated by  two professional linguists.
Table \ref{tab:stats} describes the corpus in terms of the frequency of specific reference properties that occur.
\begin{table}[h!]
    \centering
     \resizebox{\columnwidth}{!}{
    \begin{tabular}{rc|rc}
         \toprule
         Statistic &  Count &  Statistic &  Count  \\
         \hline
         Documents &  101 &
         Tokens & 29,174 \\ 
         Utterances & 3,343 & 
         Actions & 5,793 \\ 
         Markables &  7,600 & 
         Discourse old &  1960 \\ 
         Bridging & 1,053 & 
         Discourse Deixis &  500 \\ 
         Plural &  24 & 
         Ambiguous &  149 \\ 
         \bottomrule
    \end{tabular} 
    }
    \caption{Corpus Statistics}
    \label{tab:stats}
\end{table}

\section{Reference in MDC-R: 
Observations}
\label{sec:obs}


The {\MDCR} was created as a resource both to study  linguistics reference in dynamically changing 3D settings and to develop models able to carry out this type of interpretation. 
We discuss in this Section how a selection of interesting linguistic phenomena in this Section end up being captured in the annotation corpus, and its suitability for modelling development in the next.  


\subsection{Dynamically changing states of the world}

The state of the world described in  the image in Figure \ref{fig:labels2} is the result of several rounds of interaction between Architect and Builder. 
Starting from a completely empty 3D world, the Architect typically instructs the Builder to build the separate components one at a time--e.g., in the dialogue we have been discussing, the Architect first asks the Builder to create the (right) pillar shown in Figure \ref{fig:mmax2:before}, before going on asking to build the left pillar, then to put a beam on top of the two pillars, before moving on building the bell. 
All of these parts can be referred using terms such as \LINGEX{the pillar} or \LINGEX{the beam}, which in our scheme end up referring to sets of blocks.

\subsection{The effect of perspective}

The Architect and the Builder do not necessarily have the same perspective. 
Thus, the type of left/right confusions observed in the \ACRO{cups} corpus \cite{dobnik2020local} can be found in the {\MDCR} as well.
One example is the following exchange in \texttt{B36-A38-C66-1524262976017}:
\textit{``if you walk to the right side of the yellow triangle''},
 \textit{``my right?''}, \textit{``Now your left, sorry''}.


\subsection{Grounding}

In addition to misunderstandings due to perspective, a range of other types of misunderstandings can be observed, which means grounding these terms often requires rounds of clarifications. 
For instance,
in Figure \ref{fig:mmax2:before}, the Architect initially uses the term \LINGEX{the green block} to refer to the pillar built by the Builder (and shown in Figure \ref{fig:mmax2:after}). 
The uncertain Builder intially interprets \LINGEX{the 8th block on top of the green block} as a reference to the block at the top of the pillar (block a7 in the picture), but then asks a clarification request.

Notice that a fully explicit annotation of this example would require keeping track of Architect and Builder's separate views of the common ground, as done e.g., in \cite{poesio-et-al:VENEX_corpus}, but this is not currently possible in {\MDCR}, so the annotators were instructed in these cases to mark the reference intended by the speaker of the utterance.

\subsection{Intensional objects}

\begin{figure}
\centering
\includegraphics[width=\columnwidth]{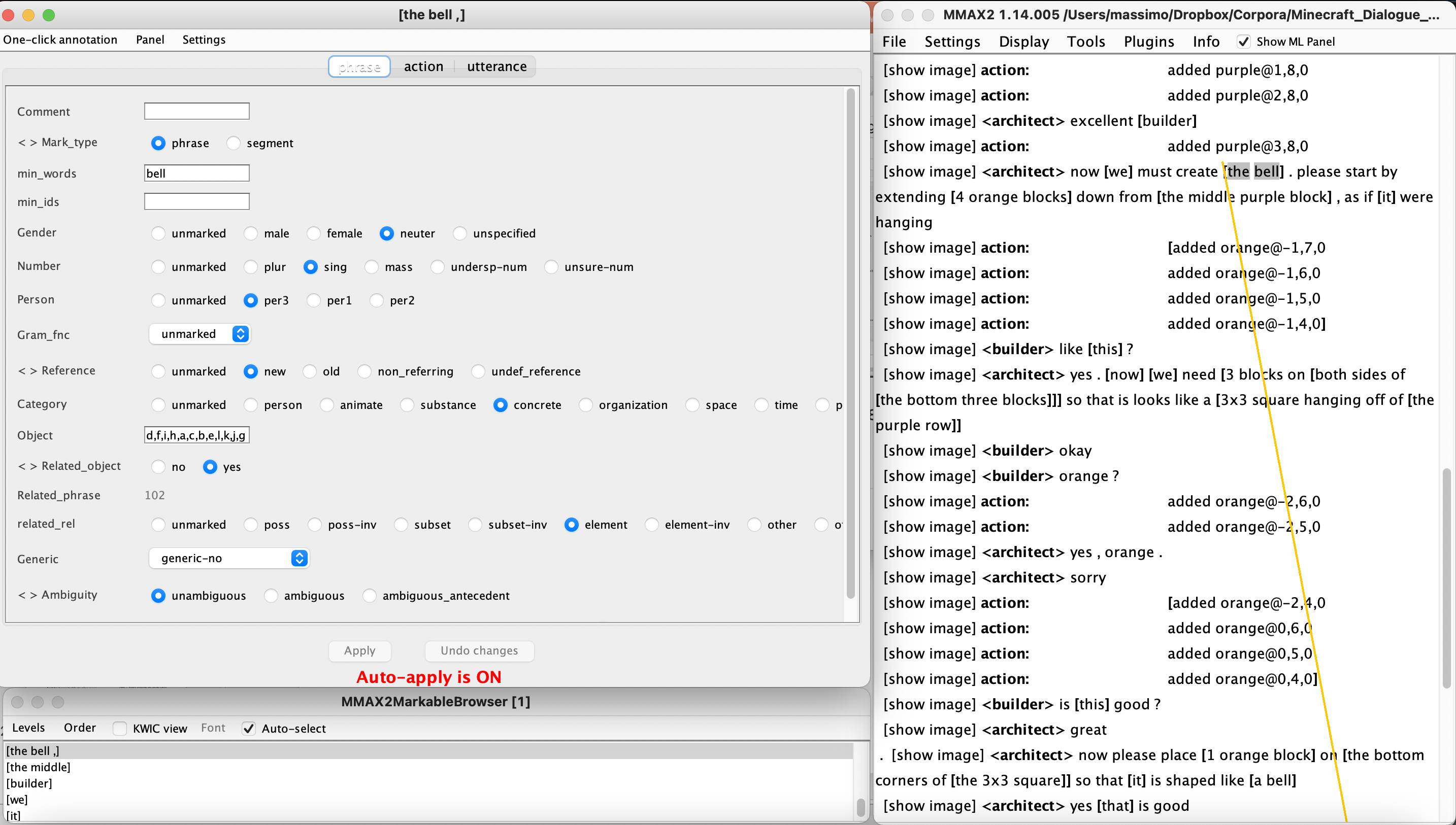}
\caption{Introducing the part of the construction to be built next, thee bell~\label{fig:intensional:before}}
\end{figure}
\begin{figure}
\centering
\includegraphics[width=\columnwidth]{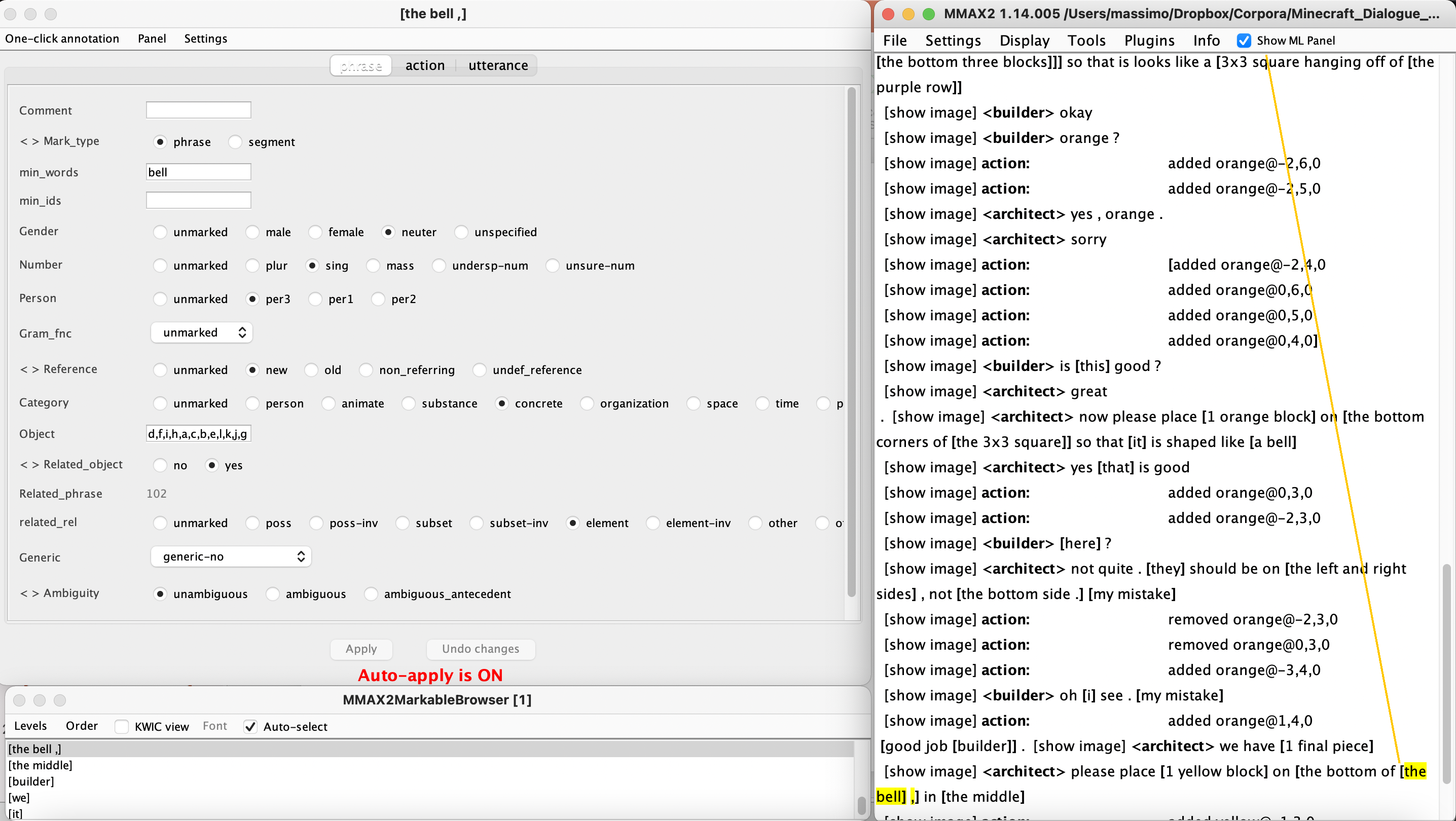}
\caption{Referring to the bell after it has been built.~\label{fig:intensional:after}}
\end{figure}

Often, the Architect starts a round of building by telling the Builder what the objective is.
For instance, in \texttt{B29-A8-C1-152286385634}, for which the target is Figure \ref{fig:labels2}, the Architect starts by telling the Builder they are going to create, \textit{``a gate with a bell all on one vertical plane''}, as illustrated in Figure \ref{fig:intensional:before}. 
This NP is not a deictic reference.  
However, after the `bell' is constructed,  the Architect's reference to \LINGEX{the bell} is treated as deictic, and as bridging 
to the first mention
(Figure \ref{fig:intensional:after}).



\section{Using the MDC-R for REC}
\label{sec:experiment}


To test the usefulness of the corpus for {\REC}, we ran experiments predicting the bounding boxes which an expression refers to in the image of the state of the world associated with that utterance. 

The primary challenge in this task is that the model must comprehend the  dialogues provided so far and infer object attributes (e.g., shape, colour, position) based on the mention, and then detect the bounding box of the object in the image, 
resulting in
cross-modal reasoning.
A further challenge arises from the dynamic and embodied environment in which the data is collected. Specifically, the perspective of the screenshots varies depending on the architect's position during gameplay
. This places a higher demand on the model to understand the voxelised world.

\subsection{Data}
To perform the {\REC} task, we construct a dataset comprising 1,150 bounding boxes for blocks, 423 referring expressions, and 101 distinct scenarios. For each scenario, we extract the mention and dialogue history up to the point of the mention, providing a textual input of the scene. The corresponding screenshot at that specific round of dialogue serves as the visual input. The ground truth label is defined as the merged bounding box encompassing all referenced blocks.

\subsection{Models}
We selected two models, Qwen2-VL~\cite{Qwen2-VL} and MDETR~\cite{kamath2021mdetr,GREC,liu2023gres}.

\textbf{Qwen2-VL}: An advanced multimodal large language model (MLLM), widely used in visual grounding, achieves state-of-the-art (SOTA) performance on RefCOCO, RefCOCO+, and RefCOCOg~\cite{kazemzadeh-etal-2014-referitgame,mao2016generation,nagaraja2016modeling}, which are widely regarded as the most classic {\REC} datasets.
%

\textbf{MDETR}: This model follows a one-stage design for {\REC}, predicting target regions directly without relying on object proposals. We adopt the implementation from \citet{GREC}, trained on the gRefCOCO~\cite{GREC,GRES} dataset, which supports multi-box prediction. To handle long referring expressions, we replace the original text encoder with Longformer~\cite{beltagy2020longformer} and fine-tune the modified model on gRefCOCO.

\subsection{Implementation Details}

We adopted two settings of the REC experiment, based on the distinct advantages of two baselines.
\textbf{Classic REC}: We used Qwen2-VL to predict a single bounding box. The input and output format template are shown on 
Table~\ref{tab:input_output_format}.
The input consists of both the system and the user perspectives, and the output is expected to contain a bounding box with a special token which could be parsed and extracted.
\textbf{GREC}: We used the MDETR model to predict a set of bounding boxes.




\subsection{Evaluation Metrics}
To evaluate the result of Qwen2-VL, we employed the \textbf{Merge-box protocol} \cite{hemanthage2024recantformer}: 
all ground-truth bounding boxes are merged into a single \textit{minimal enclosing bounding box}. This merged bounding box serves as a unified ground truth label for all blocks in one image.
To assess model performance, we compute the \textbf{Intersection over Union (IoU)} between the predicted bounding box and the merged ground-truth bounding box. Additionally, we report \textbf{Accuracy@0.25}, \textbf{Accuracy@0.5}, and \textbf{Accuracy@0.75}, where a prediction is considered correct if its IoU exceeds thresholds of 0.25, 0.5, or 0.75, respectively. 

To evaluate the performance of MDETR~\cite{GREC}, we report the mean \textbf{F1 score}
\ref{eq:f1_formula} 
at IoU thresholds of 0.25, 0.5, and 0.75.
Specifically, given a set of predicted bounding boxes and ground-truth bounding boxes, a prediction is considered a true positive (TP) if it matches (IoU $\geq$ Threshold). Predicted bounding boxes that do not match any ground-truth box are counted as false positives (FP), while ground-truth boxes without any matching prediction are treated as false negatives (FN).
\begin{equation}
\textstyle \text{F1} = \frac{2TP}{2TP + FP + FN}
\label{eq:f1_formula}
\end{equation}

In addition, for both GREC and classic REC, we calculate the \textbf{Mean IoU (mIoU)}, which represents the average IoU across all predictions in the dataset, providing a more comprehensive evaluation of overall performance.



\subsection{Results}
The results are given in Table~\ref{tab:evaluation_metrics}. While performing much better than random,\footnote{As a sanity check we tried random bounding box limits (mIoU = 5.1) and limits over the entire image (mIoU = 10.8).}  we can see that the baseline model exhibits much lower results on this dataset than on other classical {\REC} data. 
These results are noticeably lower than those typically reported in standard {\REC} datasets~\cite{kazemzadeh-etal-2014-referitgame,mao2016generation,nagaraja2016modeling} reported by~\citet{Qwen2-VL}, underscoring the challenge of accurately predicting the target bounding box in Minecraft world screenshots using dialogue context and referring expressions.

\begin{table}[H]
    \centering
    \resizebox{0.9\columnwidth}{!}{
    \begin{tabular}{lcccc}
        \toprule
        \multicolumn{4}{c}{\textbf{Metric}}\\
        
        \textbf{mIoU (\%)} & \textbf{Acc@0.25 (\%)} & \textbf{Acc@0.5 (\%)} & \textbf{Acc@0.75 (\%)} \\
        \midrule
       
        30.4 & 38.7 & 28.8 & 21.2 \\
        \bottomrule
    \end{tabular}
    }
    \caption{Evaluation of Qwen2-VL performance, where Acc denotes Accuracy under Classic REC.}
    \label{tab:evaluation_metrics}
\end{table}

\begin{table}[H]
    \centering
    \resizebox{0.9\columnwidth}{!}{
    \begin{tabular}{lcccc}
        \toprule
        \multicolumn{4}{c}{\textbf{Metric}}\\
        
        \textbf{mIoU (\%)} & \textbf{F1@0.25 (\%)} & \textbf{F1@0.5 (\%)} & \textbf{F1@0.75 (\%)} \\
        \midrule
       
        19.6 & 19.8 & 9 & 2.1 \\
        \bottomrule
    \end{tabular}
    }
    \caption{F1 performance on MDETR under GREC.}
    \label{tab:evaluation_metrics_GREC}
\end{table}
Table~\ref{tab:evaluation_metrics_GREC} shows that MDETR performs poorly under the GREC setting. We hypothesized two main causes: (1) the domain gap in long-form dialogue and virtual Minecraft scenes poses a significant challenge due to the lack of aligned data; (2) the model struggles with small object localization, as indicated by the sharp drop in F1 scores at higher IoU thresholds. A detailed example comparing the two evaluation settings is provided in Appendix~\ref{sec:appendix:two_settings_example}.

Furthermore, Table~\ref{tab:NP} provides a detailed analysis of Qwen2-VL's performance across noun phrase (NP) categories, where each NP(n) denotes a referring expression for n block(s) referenced in one Minecraft world screenshot.
The table shows that instances with smaller NP categories (i.e., NP1 to NP5) constitute the majority of the corpus.

\begin{table}[H]
    \centering
    \resizebox{\columnwidth}{!}{
    \begin{tabular}{llll}
        \hline
        \textbf{NP Category} & \textbf{mIoU (\%)} & \textbf{Accuracy@0.5 (\%)} & \textbf{Quantity} \\
        \hline
        NP1  & 21.8  & 20.3  & 246 \\
        NP2  & 27.6  & 22.7  & 66  \\
        NP3  & 39.5  & 36.4  & 33  \\
        NP4  & 36.3  & 31.8  & 22  \\
        NP5  & 19.8  & 22.2  & 9   \\
        NP6  & 76.7  & 85.7  & 7   \\
        NP7  & 20.6  & 0.0   & 1   \\
        NP8  & 49.2  & 53.8  & 13  \\
        NP9  & 41.5  & 42.9  & 7   \\
        NP10 & 71.2  & 100.0 & 1   \\
        NP12 & 95.0  & 100.0 & 2   \\
        NP13 & 4.8   & 0.0   & 2   \\
        NP14 & 0.0   & 0.0   & 1   \\
        NP15 & 27.4  & 0.0   & 1   \\
        NP16 & 98.5  & 100.0 & 4   \\
        NP18 & 65.8  & 100.0 & 2   \\
        NP19 & 96.8  & 100.0 & 1   \\
        NP20 & 98.4  & 100.0 & 3   \\
        NP28 & 95.8  & 100.0 & 2   \\
        \hline
    \end{tabular}
    }
    \caption{mIoU scores and quantities for noun phrase (NP) analysis, where NP(n) represents one mention referring to n block(s) in the Minecraft world.}
    \label{tab:NP}
\end{table}

A counter-intuitive observation is that the model performs poorly on cases involving fewer blocks (NP$<$6) while achieving higher mIoU scores for cases with more blocks. One possible explanation is that bounding boxes covering a larger area (associated with higher NP values) may inherently yield a higher overlap, thus inflating the mIoU metric. This discrepancy highlights a limitation in the current evaluation framework.

In our corpus, each block is annotated with a unique ID and precise coordinates. Although these detailed annotations provide the potential to enhance evaluation accuracy, they simultaneously introduce additional challenges for model design.

In summary, the results emphasize the complexity of the {\REC} task within the embodied visual context of Minecraft and the textual context of multi-round dialogue. Furthermore, our corpus motivates further exploration into refined evaluation methodologies and model architectures.

\subsection{Error analysis}
Manual inspection of the top 50 best- and worst-scoring examples from the results of Qwen2-VL revealed interesting patterns. Please see the Appendix for image examples.

\paragraph{Good performance} occurs when: (1) the number of blocks in the scene is low, (2) the object consists of many blocks (covering a large part of the structure), (3) object is in foreground, (4) object is clearly separated from the rest of the scene in terms of colour/position (shadows or interleaving the object with other objects is detrimental). 

\paragraph{Bad performance} occurs when (1) there are many of distractor blocks that are not part of the object (both the dialog and scene are complex) (2) object is obscured fully or partially by either the edges of the scene or by other blocks, (3) the perspective of the image makes judging distances (depth perception) difficult. These observations are in line with expectations and illustrate the challenging nature of this dataset.

\paragraph{Quantifying difficulty}
The two sets of factors above correspond to situations where the ambiguity and complexity in the image is low or high, respectively. We aimed to expand this qualitative analysis by more explicitly quantifying aspects of the data that might influence performance. To this end, we calculate pearson's $\rho$ between the model performance and different data aspects (on N1 - N10, to avoid outliers). Results, provided in Table~\ref{tbl:corrs}, confirm the findings the qualitative analysis, indicating that complexity and ambiguity, especially in terms of many distractor blocks and smaller referenced objects pose significant challenges. Furthermore, we reveal that dialog-related aspects also play a role, with longer dialogs leading to more challenging cases by producing more complex reference chains and repair structures.

\begin{table}[H]
    \centering
    {
    \resizebox{\columnwidth}{!}{
    \begin{tabular}{llll}
        \hline
        Num. object blocks (OB) &  0.21 & Num. utterances  &  -0.25   \\
        Num. scene blocks (SB)  &  -0.29  &  Num. architect utterances  &  -0.23 \\ 
        Object/Scene ratio (OB/SB)   &  0.46 & Num builder uterrances  &  -0.23   \\
        Num. distractor blocks (SB - OB) & -0.34  &   Num block remove actions  &  -0.17 \\
        Gold bounding-box area &  0.32  & Num. block add actions  &  -0.26 \\
    \hline
    \end{tabular}
    }    
    \caption{Pearson's $\rho$ between IoU and data aspects.}
    \label{tbl:corrs}
    }
\end{table}

\section{Conclusion} 




We introduced the Minecraft Dialogue Corpus with Reference, a multimodal dialogue dataset that adds deictic and anaphoric reference annotations to it.  
The new version of the {\MDC} provides both anaphoric and deictic reference labels for objects in the scenes,  ids for the blocks, bounding boxes for the objects referred to, 
and a dynamic setting in which the actions of the speakers modify the world state (as well as speaker perspectives on the scene) as the dialog progresses. 
Moreover, {\MDCR} is the first  dataset which provides information about the constituent parts of the referenced objects. As such, the dataset can be used for a wide variety of diverse tasks and will constitute both a valuable tool for studying collaborative dialogue as well as a very challenging benchmark for the reasoning abilities of next-generation multi-modal LLMs.

\section*{Limitations}

The key limitation of this work is that we were only able to annotate part of the original {\MDC} in the time available. 
We hope to carry out more annotations in the future. 
A second limitation is that only one perspective is annotated, that of the speaker--as discussed in the paper, in some cases it would be good to explicitly annotate the different points of view of Architect and Builder. 
We also hope to address this limitation in future work. 

\section*{Acknowledgements}

This research was funded by ARCIDUCA, EPSRC EP/W001632/1

\bibliography{custom}


\section{Appendix}
\label{sec:appendix}
\subsection{Prompt}

\begin{table}[H]
    \centering
    \small
    \begin{tabular}{p{1cm} p{5.8cm}}  
        \hline
        \multicolumn{2}{c}{\textbf{Input format}}\\
        \hline
        System& You are a helpful agent to understand a mention in the last sentence of a dialogue in a Minecraft scenario, and detect the bounding boxes of the target block(s). 
        
        <|dialogue$\_$start|><|dialogue$\_$end|> is the dialogue. 
        
        <|mention$\_$start|><|mention$\_$end|> is the mention.
        
        You need to output the bounding box in <|box$\_$start|><|box$\_$end|>. \\
        \hline
        \multirow{2}{*}{User}& <|vision$\_$start|>picture<|vision$\_$end|>\\
        & <|dialogue$\_$start|>dialogue<|dialogue$\_$end|>
        detect the bounding box of <|mention$\_$start|>mention<|mention$\_$end|>.\\
        \hline
        \multicolumn{2}{c}{\textbf{Output format}}\\
        \hline
        Agent& ...<|box$\_$start|>bounding$\_$box<|box$\_$end|>\\
        \hline
    \end{tabular}
    
    \caption{The input and output formats. In the input, an additional instruction from the system perspective is added to guide the model on how to understand the task.}
    \label{tab:input_output_format}
\end{table}

\subsection{Example for two distinct evaluation settings} 
\label{sec:appendix:two_settings_example}
\begin{figure}[H]
    \centering
    \includegraphics[width=0.9\linewidth]{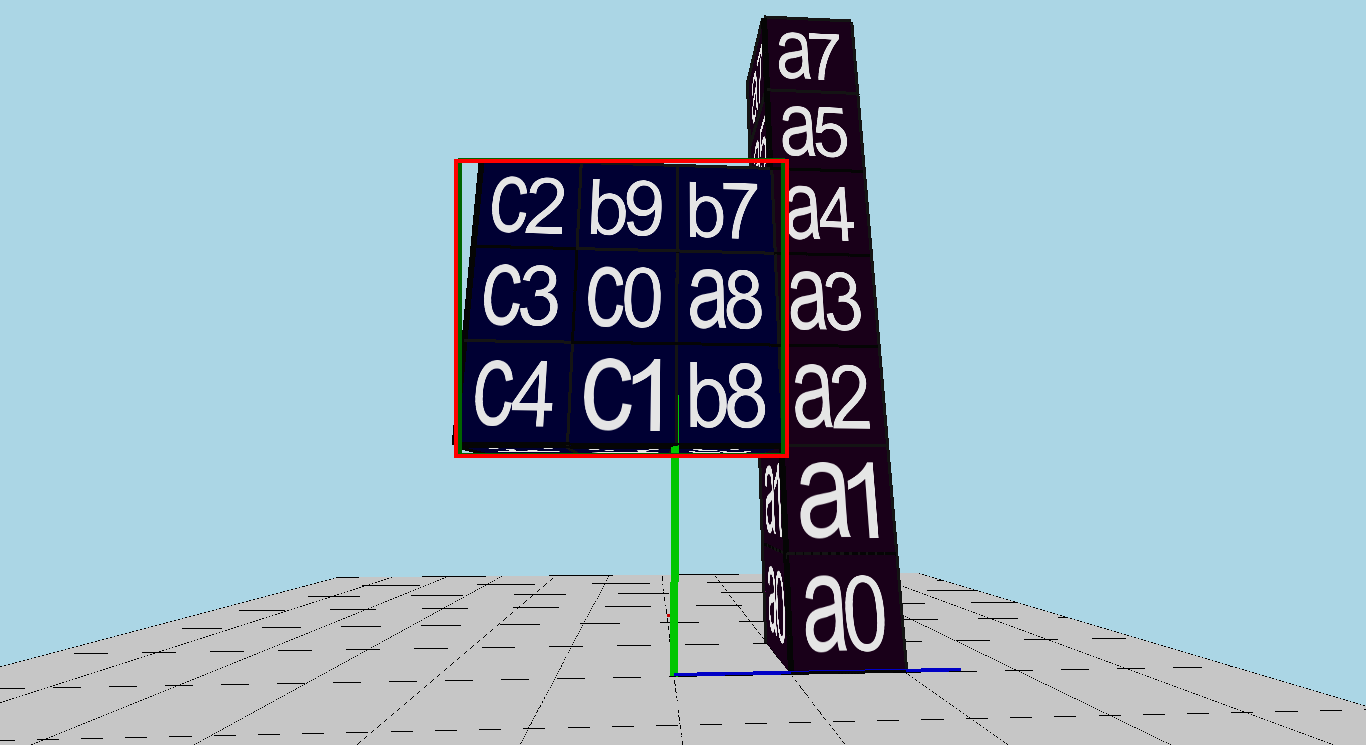}
    \caption{Example case of recognizing merged blocks by Qwen2-VL under the REC setting.}
    \label{fig:exampleREC}
\end{figure}

\begin{figure}[H]
    \centering
    \includegraphics[width=0.9\linewidth]{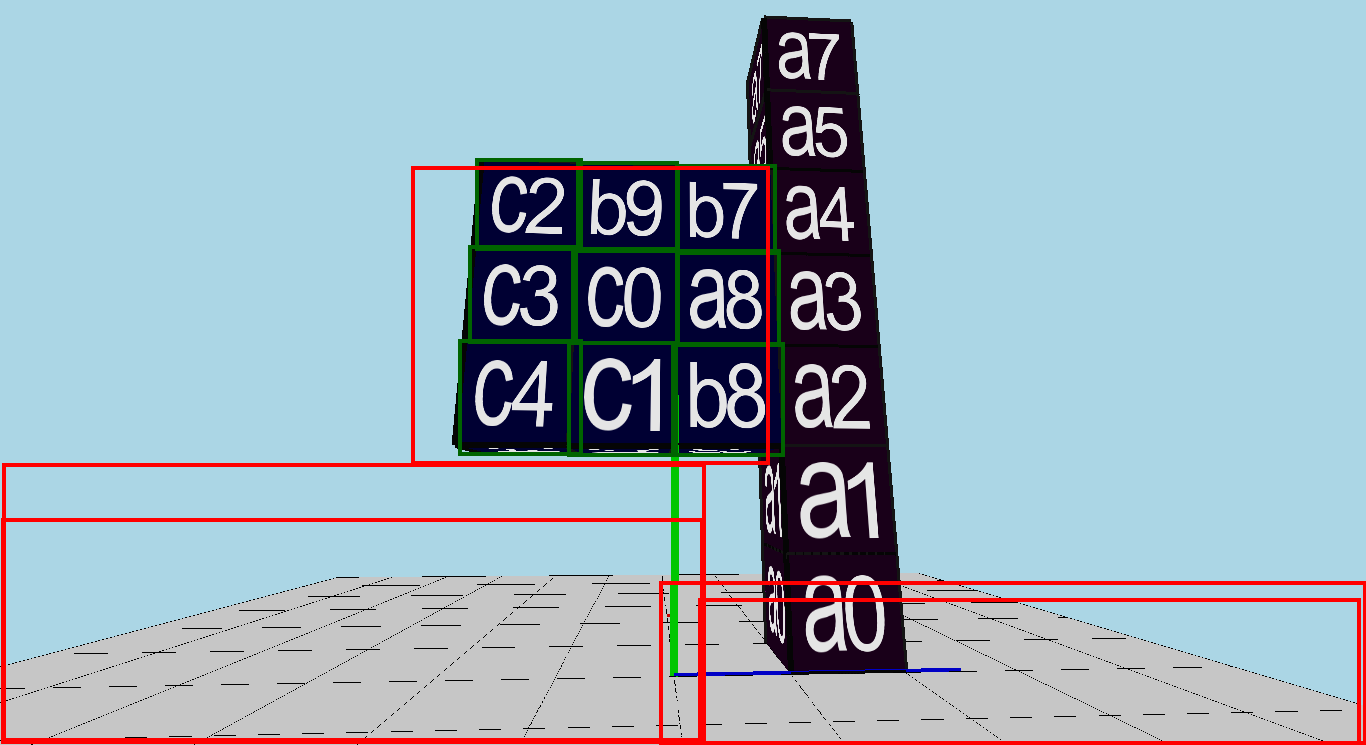}
    \caption{Example case of recognizing independent blocks by MDETR under the GREC setting.}
    \label{fig:exampleGREC}
\end{figure}

We present model predictions under two different evaluation settings. The gold bounding box is shown in green, and the predicted one in red. As shown in Figure~\ref{fig:exampleREC}, under the REC setting, the model easily recognizes the large 3×3 blue square, as the target occupies a prominent area in the image. In contrast, as illustrated in Figure~\ref{fig:exampleGREC}, the GREC setting requires the model to precisely locate each individual block that forms the square. This significantly increases the difficulty, demanding a deeper understanding of the expression and a precise reference to the dialogue to consider all mentioned blocks.

\subsection{Example model outputs}
\begin{figure}[H]
    \centering
    \includegraphics[width=0.9\linewidth]{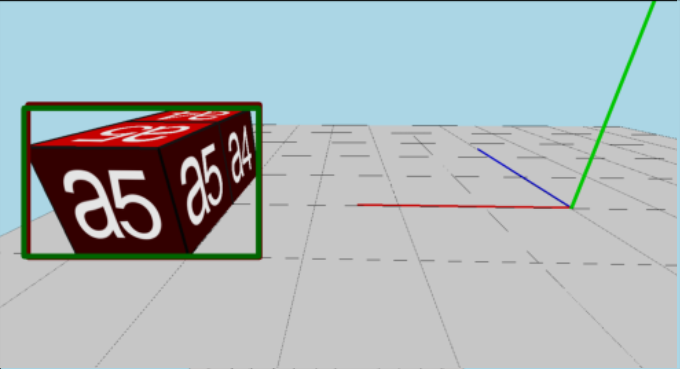}
    \caption{Example case where there are only a few blocks in the scene, making the task less challenging.}
    \label{fig:exampleA1}
\end{figure}

\begin{figure}[H]
    \centering
    \includegraphics[width=0.9\linewidth]{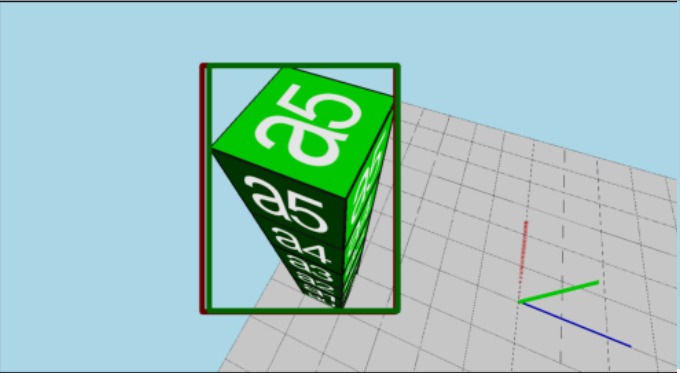}
    \caption{Example of a case where the object covers a significant part of the structure (often the whole structure). This makes the example less challenging even if there are many blocks involved.}
    \label{fig:exampleA2}
\end{figure}

\begin{figure}[H]
    \centering
    \includegraphics[width=0.9\linewidth]{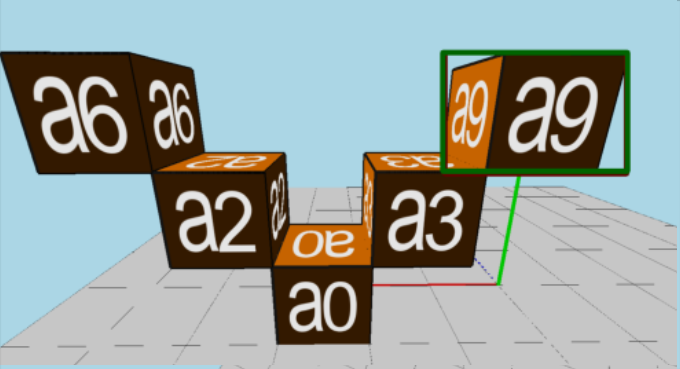}
    \caption{Example case where the object block is in the foreground and easy to see.}
    \label{fig:exampleA3}
\end{figure}

\begin{figure}[H]
    \centering
    \includegraphics[width=0.9\linewidth]{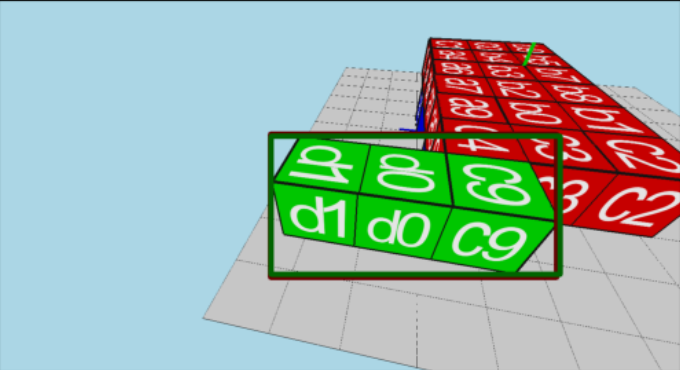}
    \caption{Example case where the object is clearly separated by colour.}
    \label{fig:exampleA4}
\end{figure}

\begin{figure}[H]
    \centering
    \includegraphics[width=0.9\linewidth]{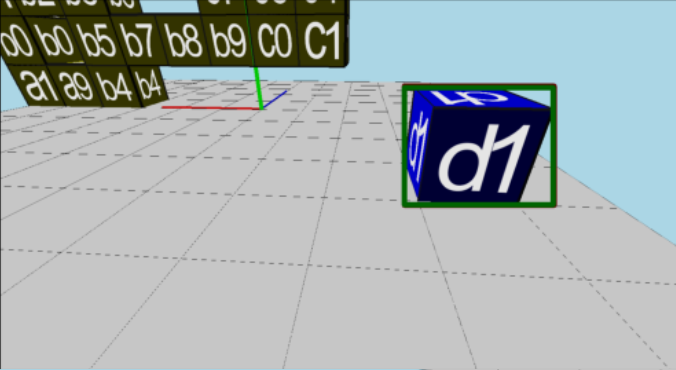}
    \caption{Example case where the object is clearly separated in terms of space.}
    \label{fig:exampleA42}
\end{figure}

\begin{figure}[H]
    \centering
    \includegraphics[width=0.9\linewidth]{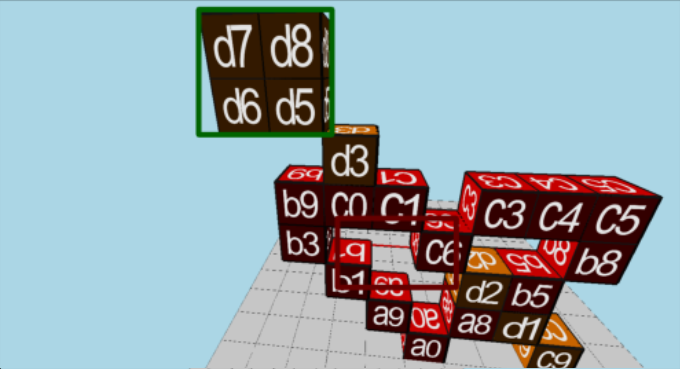}
    \caption{Example case where there is a large number of distractor blocks which do not belong to the referenced object. These cases are extremely challenging.}
    \label{fig:exampleB1}
\end{figure}

\begin{figure}[H]
    \centering
    \includegraphics[width=0.9\linewidth]{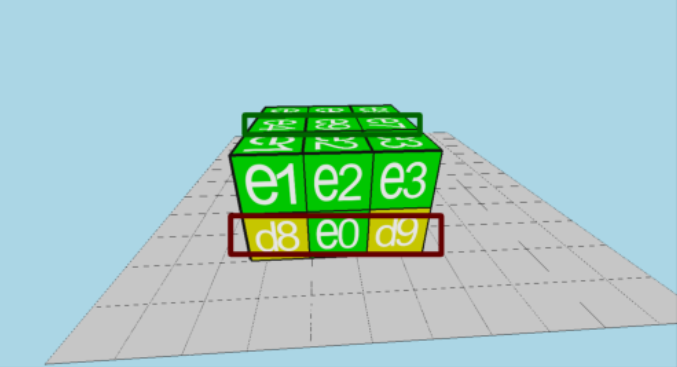}
    \caption{Example case where the object is partially obscured by. These are often very challenging.}
    \label{fig:exampleB2}
\end{figure}

\begin{figure}
    \centering
    \includegraphics[width=0.9\linewidth]{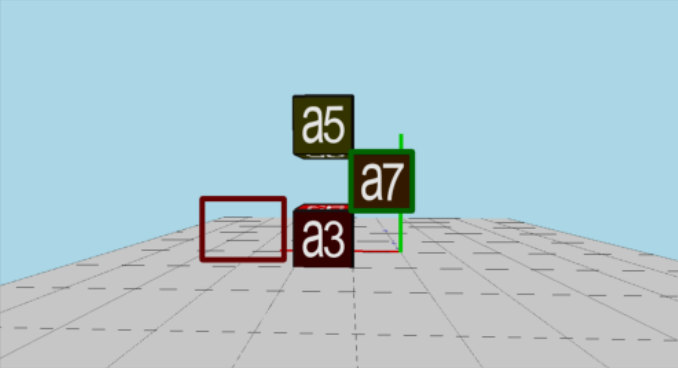}
    \caption{Example case where the perspective makes depth perception very difficult, making the task harder.}
    \label{fig:exampleB3}
\end{figure}

This section provides images of the model outputs for particularly easy (Figure~\ref{fig:exampleA1}, Figure~\ref{fig:exampleA2}, Figure~\ref{fig:exampleA3}, Figure~\ref{fig:exampleA4}, and Figure~\ref{fig:exampleA42}) or difficult (Figure~\ref{fig:exampleB1}, Figure~\ref{fig:exampleB2}, Figure~\ref{fig:exampleB3}) cases. The gold bounding box is green, while the predicted one is red.


\end{document}